\documentclass[11pt,a4paper]{article}
\usepackage[hyperref]{emnlp2020}
\usepackage{times}
\usepackage{latexsym}
\usepackage{amsmath}
\usepackage{amsfonts,amssymb}
\usepackage{bbm}
\usepackage{color}
\usepackage{algorithm,algorithmic}
\usepackage{subcaption}
\usepackage{graphicx}
\usepackage{booktabs}
\usepackage{diagbox}
\usepackage{multirow}
\usepackage{longtable}
\usepackage{extarrows}
\usepackage{appendix}

\newcommand*\samethanks[1][\value{footnote}]{\footnotemark[#1]}

\usepackage{microtype}

\title{Towards Collaborative Question Answering: A Preliminary Study}

\author{Xiangkun Hu\textsuperscript{1,}\thanks{\ \  Equal contribution.} , Hang Yan\textsuperscript{2,}\samethanks \ , Qipeng Guo\textsuperscript{1} , Xipeng Qiu\textsuperscript{2}, Weinan Zhang\textsuperscript{3}, Zheng Zhang\textsuperscript{1}\\
  \textsuperscript{1}AWS Shanghai AI Lab \\
  \textsuperscript{2}School of Computer Science, Fudan University \\
  \textsuperscript{3}Shanghai Jiao Tong University \\
  \texttt{\{xiangkhu, gqipeng, zhaz\}@amazon.com} \\
  \texttt{\{hyan19,xpqiu\}@fudan.edu.cn}\\
  \texttt{wnzhang@apex.sjtu.edu.cn}\\}

\date{}

\begin{document}
\maketitle
\begin{abstract}

Knowledge and expertise in the real-world can be disjointedly owned. To solve a complex question, collaboration among experts is often called for. In this paper, we propose CollabQA, a novel QA task in which several expert agents coordinated by a moderator work together to answer questions that cannot be answered with any single agent alone. We make a synthetic dataset of a large knowledge graph that can be distributed to experts. We define the process to form a complex question from ground truth reasoning path, neural network agent models that can learn to solve the task, and evaluation metrics to check the performance. We show that the problem can be challenging without introducing prior of the collaboration structure, unless experts are perfect and uniform. Based on this experience, we elaborate extensions needed to approach collaboration tasks in real-world settings.

%Knowledge and expertise in real-world can be disjointedly owned, and not all of them are digitized. In this paper, we study the problem where multiple hybrid agents must collaborate to respond to a question. In order to cope with the fact that some agents are human agents, we impose the constraint that the questions posted among the agents must be well-formed natural sentences. We propose CollabQA, a novel QA task in which several QA agents need to collaborate to answer questions that cannot be answered with a single agent. Each agent has the partial and incomplete knowledge for a question. To perform the task well, an agent not only needs to infer what it knows but also what it does not, and formulate follow-up question(s) to other agents when necessary. We propose a simple solution to solve the problem under reinforcement learning framework. We explore the feasibility of CollabQA and the effectiveness of our approach on the setting where four agents collaborate to answer complex questions. Results on a synthetic dataset show that our method effectively trained the agents to collaborate to answer questions with high accuracy and good conversation quality.
\end{abstract}

\section{Introduction}

One of the fascinating aspects of human activity is collaboration: despite the limitations of our individual experience and knowledge, we can collaborate to solve a problem too challenging for any one person alone. In the context of this paper, we are interested in collaboration via rounds of questions and answers \emph{internal} to a panel of experts responding to an \emph{external} question. Forms of such activities have broadened into the realm of robots as well. For instance, customer service is automated with the backing of machine agents, each holding expert knowledge in a specific domain.

\begin{figure}[t]
    \centering
    \includegraphics[scale=0.5]{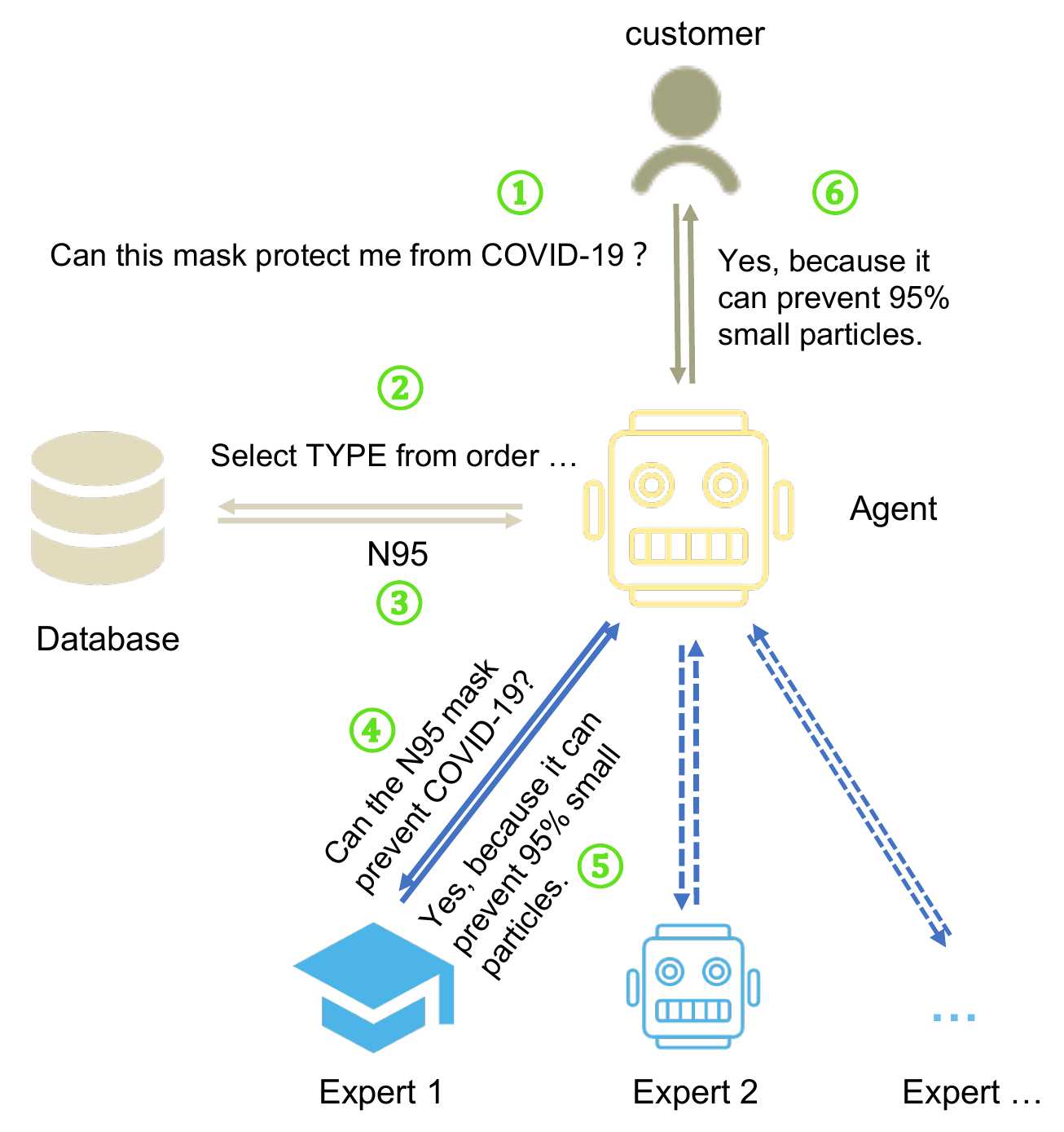}
    \caption{An example in a hypothetical customer service scenario. The customer asks a question about a feature of the product in the order he/she is about to place. However the database of the service agent doesn't contain the information. Instead of responding with something like ``\textit{Sorry I don’t know}”, the better way is to get help from human experts, or other QA agents.}
    \label{fig:motivation}
\end{figure}

Figure~\ref{fig:motivation} shows a hypothetical customer service example, where an AI agent is serving a customer who is about to place an order of a mask. Even though the agent has access to the features (e.g., `N95') of the mask in its local database, it cannot answer the question ``\textit{Can this mask protect me from COVID-19?}''. Instead of responding with ``\textit{Sorry I don't know.}'' as most of the current QA systems do, it can reroute a new question ``\textit{Can the N95 mask prevent the COVID-19?}" to a human expert.

We call this task \emph{CollabQA} where a single agent (human or robot) cannot reason and respond to a complex question, but collectively they can. In other words, knowledge is not shared across agents, but the union contains the required reasoning path, which necessitates collaboration.

To solve the problem in its general form is hard. In this paper, we take a few steps forward by proposing 1) a simplified version of CollabQA task where one front-serving agent decomposes an external question into simple ones for the rest of the experts to answer, 2) a synthetic dataset of a large knowledge graph that can be distributed to experts, and 3) a set of baseline models and the associated evaluation metrics.

Despite this very simple form, we show the problem can be very challenging.
Our overall conclusion is that, even with such a simple setting where 1) knowledge is clearly decomposed, 2) collaboration is passive, and 3) questions and answers are formed with simple templates and node prediction, training a good collaboration policy remains challenging, unless we add a strong prior reflecting the collaboration structure, and assume collaborators that are both perfect and uniform. We use these experiences to reflect how to improve this task to gradually approach collaboration tasks with more real-world flavor.

% Our experiences motivates [new dataset and task discussion] Blah blah blah Blah blah blah Blah blah blah Blah blah blah Blah blah blah Blah blah blah Blah blah blah Blah blah blah Blah blah blah Blah blah blah Blah blah blah Blah blah blah Blah blah blah Blah blah blah and share our lessons.
%These baselines differ by strength of labels and by whether Q\&A must be in forms of natural languages or not, the latter requirement stems from the fact that a future deployment of CollabQA will involve both human and machine experts.

The rest of the paper is organized as follows.  Section~\ref{secton:task_setting} formally defines the CollabQA task setting and shows the toy dataset we synthesized for a preliminary study. Section~\ref{section:approach} describes the approach we proposed to for the task. We show experimental results and their enlightenment in Section~\ref{section:experiment}. Section~\ref{section:related_work} surveys related works to CollabQA and discuss the key differences. And Section~\ref{section:discussion} discusses some potential directions for future works.

\section{Opening Remarks}
This paper was initially submitted to the EMNLP 2020 on June 3, 2020. The reviewers' primary concern about this paper was that it lacked real data experiments and rejected this paper. Since then, we thought we might have time to polish this paper further, but the research direction of our team changed to other fields. Therefore, we did not have the chance to go deeper in this direction. Some of the settings or discussions may be interesting to the community, so we decided to release this paper on the arxiv. Since 2020, we have noticed more related papers in this section. We list some of them for the readers' reference and leave other parts of this paper almost unaltered to its initial version.

To make agents collaborate, we usually need to decompose a complex task into simpler ones so that different agents can tackle these simple tasks. \citet{DBLP:journals/tacl/WolfsonGGGGDB20} defined several operators, and a complex question will be decomposed into several sub-queries so that each sub-query will have only one operator. Based on this principle, \citet{DBLP:journals/tacl/WolfsonGGGGDB20} annotated a large dataset BREAK which can be served as a good starting point for Question Answering (QA) collaboration. \citet{DBLP:conf/acl/HeBEL17} proposed a dataset that requires two people, each with a distinct private list of friends, to find their mutual friends through talking. CEREALBAR proposed in \cite{DBLP:conf/emnlp/SuhrYSYKMZA19} is a collaborative game, which requires an instructor and a follower to collaborate to gather three cards in a virtual environment. The instructor can use natural language to pass messages to a follower, but not vice versa. The instructor has to learn to use better instructions to achieve better scores. \citet{DBLP:conf/naacl/KhotKRCS21} proposed to use natural language to make several existing QA models collaborate so that they together can solve a question that cannot be solved solely by any existing QA models. And they further proposed a synthetic 
benchmark COMMAQA which can facilitate the research of collaboration QA \cite{DBLP:journals/corr/abs-2110-08542}.

\section{The CollabQA Task}
\label{secton:task_setting}

\subsection{Notations and Settings}

\begin{table}[t]
    \centering
    \begin{tabular}{l|l}
\hline
Notations         &  Description \\
\hline
$P_i$             & The $i$-th panelist. \\
$Q$               & The external complex question. \\
$q^{t}$           & Utterance by $P_0$ at $t$-th dialog turn. \\
$u_i^{(t)}$       & Response of $P_i$ at $t$-th dialog turn. \\
$\text{KG}_i$     & Knowledge graph owned by $P_i$. \\
$\tau(Q)$         & The reasoning path of $Q$. \\
$T$               & Dialog turns. \\
\hline
    \end{tabular}
    \caption{Notations of CollabQA.}
    \label{tab:notations}
\end{table}

The general setting of CollabQA simulates a group of panelists $\{P_i\}_{i=0}^{n}$, out of which $P_0$ is a special: it is the front-serving \emph{receptionist} and the representative to the external world, it is also the \emph{moderator} of the collaboration among $\{P_i\}_{i=1}^{n}$, who we term as the panelists. When $P_0$ receives an external question $Q$, it broadcasts an utterance $q^{(1)}$ to the panelists and collects responses $\{u_i^{(1)}\}_{i=1}^n$ from them. This process continues iteratively, each round is a tuple $(q^{(t)}, \{u_i^{(t)}\}_{i=1}^n)$, until a maximum of $T$ turns, and/or when $P_0$ is able to generate the final response which includes ``UNK'', that means ``I don't know''. Notations used in this paper are listed in Table~\ref{tab:notations}.

The panelists $\{P_i\}_{i=1}^{n}$ owns a list of knowledge graphs, $\text{KG}_1, \text{KG}_2, \dots, \text{KG}_n$, and their union $\text{KG} = \cup_{i=1}^n \text{KG}_i$ is the total graph. Questions are usually complex in the sense that they cannot be answered by one single agent. However, they are always answerable by $\text{KG}$. In other words, $\tau(Q)$, the reasoning path of question $Q$ can cut across different graphs but is always contained within $\text{KG}$. As such, $P_0$ must generate multiple polls to the panelists to stitch together $\tau(Q)$. Our objective is to minimize the total number of turns while maximizing the success rate.

\subsection{A Toy Task}
%Given that we want to simulate a QA system to answer up to $n$ turns, $G$ must contain reasoning paths of at least $n$ hops. Each hop is a tuple of $(e_i, r_{ij}, e_j) \in G$. Furthermore, $G$ should be such that it can be decomposed into $n$ sub graphs %of roughly equal size and  with some domain affinity.
%For this purpose, we synthesize a knowledge graph which stores knowledge about a list of fabricated \emph{person}, \emph{company} and \emph{city} entities as well as their relationships. We split the whole graph into three parts, named $G_1$, $G_2$ and $G_3$. $G_1$ contains the knowledge of persons, $G_2$ stores the knowledge of companies and $G_3$ is about cities. The size of the graph can be arbitrarily large, in our experiment, we use 20,000 persons, 1,000 companies and 100 cities. The details of the KGs are listed in Appendix~\ref{appendix:kg_details}.
Inspired by the bAbI task~\cite{weston2015towards}, we construct a CollabQA dataset, which contains a series of QA pairs and 3 supporting knowledge graphs.

We first construct $\text{KG}_1, \text{KG}_2, \text{KG}_3$ consisting of fabricated \emph{person}, \emph{company} and \emph{city} entities and their relations. They stores the knowledge of $N_1$ persons, $N_2$ companies and $N_3$ cities respectively. The details of the three knowledge graphs are listed in Appendix~\ref{appendix:kg_details}. They are assigned to the panelists $P_1$, $P_2$ and $P_3$ respectively as their knowledge.

\begin{figure}[h]
    \centering
    \includegraphics[scale=0.42]{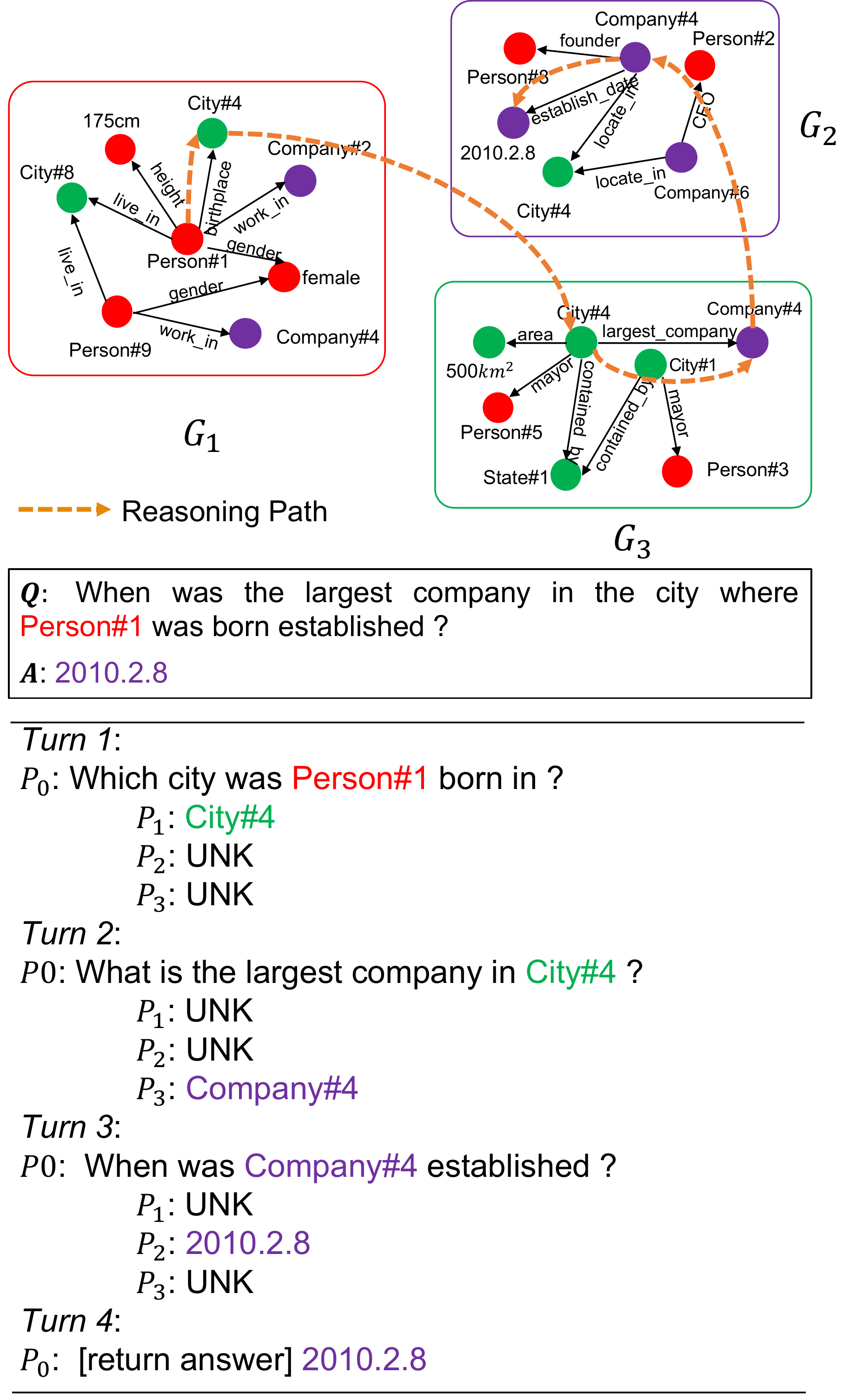}
    \caption{Illustration of Toy CollabQA Task: an example of QA pair and the ideal collaborative process.}
    \label{fig:reasoning_path}
\end{figure}

Then we synthesize QA pairs from the knowledge graphs as well as the reasoning paths. Each question needs a cross-graph multi-hop reasoning.
To illustrate the process of creating the dataset examples, we use an example to show how to create a 2-hop question: from a node ``Person\#1'' in $\text{KG}_1$, we follow a path with many-to-one or one-to-one types of relations, for example the ``birthplace'' relation and get a triplet (Person\#1, birthplace, City\#4); then we start from node ``City\#4'' in $\text{KG}_3$ to search a triplet (City\#4, largest\_company, Company\#4). Then we combine the two triplets into a reasoning path:
\begin{align}
    \text{Person\#1} & \xlongrightarrow{\text{birthplace}} \text{City\#4} \notag \\
    &\xlongrightarrow{\text{largest\_company}} \text{Company\#4}
\label{eq:reasoning_path_example}
\end{align}
so the final answer is the entity Company\#4 in the end of the path. The question $Q$ asking about Company\#4 following the reasoning path is ``What is the largest company in the city where Person\#1 was born?'', which is generated by templates.
A more complex example is shown in the upper part of Figure~\ref{fig:reasoning_path}.

The reason we use many-to-one or one-to-one types of relations during the search, is that it ensures the entities occurred in the path are unique, so that we can decompose the question into sub questions and each with unique answer. In general, to generate an $n$-hop question, we randomly pick an entity node and perform $n$-hop Depth First Search (DFS). Note that multiple edges may exist between a pair of entities (as in a person may live and die in the same city).

We limit the communications among the panelists are natural language. Therefore, $P_0$ needs to learn how to ask questions. To alleviate the burden of text generation, we pre-define a set of templates of sub questions. So, for $\tau(Q)$ in equation \ref{eq:reasoning_path_example}, the sub questions are ``Which city was Person\# born in ?'' for the first hop and ``What is the largest company in City\#4 ?'' for the second hop.  The bottom part of Figure~\ref{fig:reasoning_path} shows an ideal collaborative process.

Table~\ref{tab:data_stat_overall} lists the overall statistics of the dataset.

\begin{table}[h]
    \centering
    \begin{tabular}{l|c}
\hline
Statistics description     & Value \\
\hline
Train set size    & 66,800     \\
Dev set size      & 8,350     \\
Test set size     & 8,350     \\
\# of templates of $Q$& 49     \\
\# of templates of simple questions & 28 \\
\hline
    \end{tabular}
    \caption{Overall statistics of the CollabQA dataset.}
    \label{tab:data_stat_overall}
\end{table}

% \paragraph{Generating QA examples} All the answers are nodes in $G$, so that we can search reasoning paths for them. As the reasoning path to the external question $Q$ cut across all three KGs, $Q$ is a 3-hop question. From a random node in the KGs, we extract a 3-hop reasoning path across all three KGs ended with another node, then the question template is created. One example of this process is shown in Figure~\ref{fig:reasoning_path}, the search of a reasoning path starts with a ``PersonName'' node, search 3 hops with Depth First Search (DFS), and ends with a ``establish date value'' node. So the corresponding question is asking about a company's establish date with 3-hop relation with a person. To ensure each intermediate node in the reasoning path has an unique value, we only consider many-to-one or one-to-one relations during searching. We found 49 types of question template at first, then filtered out templates that occurred less than 1,000 times in the dataset. Finally we have 11 templates of $Q$. All types of reasoning paths and their corresponding question templates used in this dataset can be found in Appendix~\ref{appendix:templates}.

\subsection{Links to other QA Tasks}
\label{section:related_work}
CollabQA can be regarded as an combination of several kinds of QA tasks: knowledge graph question answering (KGQA), multi-hop QA, multi-turn dialogue.
%where each panelist privately owns a partial knowledge graph.

\paragraph{KGQA} In CollabQA, each panelist is a KGQA system. KGQA assumes that each panelist can answer the questions according its own KG, though it may need one- or several-step reasoning.

\paragraph{Multi-hop QA} In multi-hop QA, the supporting facts of a question are scattered in different sources. Most models for multi-hop QA assume they can access all the sources. Different from multi-hop QA, the supporting facts in CollabQA are separately owned by different panelists. Each supporting fact is not accessible except its owner. Therefore, panelists need communicate with each other to exchange information.

\paragraph{Multi-turn Dialogue} The multi-turn dialogue usually occurs between human and agent. CollabQA aims to develop multi-turn interactions among several agents (panelists).

As such, CollabQA is more challenging than KGQA, multi-hop QA and multi-turn dialogue.

\section{Proposed Approach}
\label{section:approach}

In our setting, panelists collaborate \emph{passively} in that they respond with what they know or else with ``UNK''. Therefore, $P_0$ leads the process of collaboration. Our general approach consists of two stages: 1) pre-train the panelists with supervised learning; 2) train the collaboration policy with reinforcement learning.

\subsection{Panelists}
Panelists share the same model architecture: a \emph{Graph Encoder} that encode the knowledge graph into a graph representation matrix $H^{(\text{KG})}$, a \emph{Question Encoder} that encodes the incoming question $q^{(t)}$ into $\mathbf{h}(q^{(t)})$, feeding both into a Node Selector then picks an entity as the answer.

\paragraph{Graph Encoder}
Without ambiguity, we call each knowledge graph owned by expert agents $\text{KG} = (\mathcal{V},\mathcal{E})$. $\text{KG}$ is a heterogeneous graph consisting of different types of entities $\mathcal{V}$ and their relations $\mathcal{E}$. The form of each relation is a triplet $\{(u, rel,v)\}$, where $u,v\in \mathcal{V}$ and $rel\in \mathcal{R}$ and $\mathcal{R}$ is the set of relation types.

We come up with a modified version of {Relational Graph Convolutional Network (R-GCN)}~\cite{schlichtkrull2018modeling} as the graph encoder. R-GCN encodes the graph by aggregating the neighbor and edge information to the nodes. Given a node $v$ in $\text{KG}$, let $\mathbf{h}^{(l)}_v$ denote its representation at the $l$-th layer of R-GCN, then
\begin{equation}\small
    \mathbf{h}^{(l+1)}_v = \delta \left( \sum_{rel\in \mathcal{R}} \sum_{u\in \mathcal{N}^{rel}_v} \frac{1}{c_{v,rel}} W^{(l)}_{rel} \mathbf{h}^{(l)}_u + W^{(l)}_0 \mathbf{h}^{(l)}_v \right),
\end{equation}
where $\delta$ is an activation function, $\mathcal{N}^{rel}_v$ denotes the neighbors of $v$ which have relation $r$ with $v$, $W^{(l)}_{rel}$ is the weight matrix of relation $rel$ at the $l$-th layer, and $\frac{1}{c_{v,rel}}$ is a normalization factor. After $L$ aggregations, the final representations of the nodes are $H^{(\text{KG})}$.

However, R-GCN suffers from high GPU memory usage, making it hard to scale to large graphs. The reason is that computing the message includes a direct tensor operation that will produce a very large tensor, espeically when number of relations is large. On the other hand, if we compute the messages with a for-loop, the speed of the aggregation will suffer. To get rid of this problem, we make modifications similar to \cite{vashishth2020compositionbased} but simpler and sufficient for this task in CollabQA dataset: each relation is modelled by a trainable vector $\mathbf{h}_{rel}$ instead of a matrix $W_{rel}$, then the aggregation process changes to:
\begin{equation}\small
    \mathbf{h}^{(l+1)}_v = \delta \left(\frac{1}{c_{v,rel}} \sum_{rel, u\in \mathcal{N}^{rel}_v
    } \mathrm{MLP}^{(l)}([\mathbf{h}^{(l)}_v, \mathbf{h}^{(l)}_{rel}, \mathbf{h}^{(l)}_u])  \right).
\end{equation}%
In our experiments, we observe significant GPU memory saving. Our implementation leverages the DGL package for its superior GPU performance ~\cite{DBLP:journals/corr/abs-1909-01315}.

\paragraph{Question Encoder}
Similarly, we call the question to the expert agents $q$. The representation of a question $q$ is computed by BiLSTM~\cite{hochreiter1997long}:
\begin{equation}
\mathbf{h}^{(q)} = \mathrm{BiLSTM}(q),
\end{equation}

\paragraph{Node Selector}
Node selector performs an attention operation with $\mathbf{h}^{(q)}$ on $H^{(\text{KG})}$, and returns attention scores $\alpha^{(\text{KG})}$ on $H^{(G)}$ as the likelihood of selecting each node as answer:
\begin{equation}
     \alpha^{(\text{KG})} = (H^{(\text{KG})})^T W \mathbf{h}^{(q)}
\end{equation}
then the answer is the value of the node which has the highest attention value.

\subsection{Moderator and Collaboration Policy}

%The model of $P_0$ is a template selection policy which takes in a dialog history and outputs a probability distribution of a set of simple question templates $U$. It consists of a history encoder that encodes dialog history $H^{(t)}$ at turn $t$ into a state representation $s_t$, and a policy $\pi(s_t)$ that computes the distribution $P(U|s_t)$.

%  $P_0$ is an agent and represents the interface between the external questionnaire and the group. When $P_0$ receives an external question $Q$, it broadcasts an utterance $q^{(1)}$ to the committee $\{P_i\}_{i=1}^{n}$ and collects responses $\{u_i^{(1)}\}_{i=1}^n$ from them.

% The \emph{receptionist} $P_0$ plays the role of receptionist and manager of the collaboration.
$P_0$ coordinates collaboration according to a learned collaboration policy. At turn $t$, $P_0$ takes $a^{(t)}$ according to its current state $\mathbf{s}^{(t)} = f_s(d^{(t)})$,
%\paragraph{State Encoder
%The state $s^{(t)}$
where $d^{(t)}$ is the dialog history up to $t$,
$d^{(t)}= [Q, q^{(1)}, \{u_i^{(1)}\}_{i=1}^n,$ $ \cdots, q^{(t-1)},\{u_i^{(t-1)}\}_{i=1}^n]$.
% , where $q^{(t)}$ is the sub question generated by $P_0$ at turn $t$ and $\{u_i^{(t)}\}_{i=1}^n$ is the responses from panelists for  $q^{(t)}$.
% We encode the dialog history $s^{(t)}$ into vector representation $\mathbf{s}^{(t)}$:
% \begin{equation}
% 	\mathbf{s}^{(t)} = f_s(s^{(t)}),
% \end{equation}
The state encoder $f_s(\cdot)$ can be any neural model; here we use BiLSTM.

%\paragraph{Action}
The action space includes asking a new sub question or returning the final answer.
To alleviate the burden of text generation, $P_0$ generates sub question by selecting templates from a predefined set $\mathcal{U}$. To enable $P_0$ to determine whether to finish the collaboration and return the answer of $Q$, we add a special template in $\mathcal{U}$  which stands for ``finish the collaboration''.
Thus, we use a simple Multilayer perceptron (MLP) to implement the collaboration policy $\pi(a^{(t)} | \mathbf{s}^{(t)})$, which
takes $\mathbf{s}^{(t)}$ as input, outputs a probability distribution over the list of templates.

In CollabQA dataset, at each dialog turn, only one of the answers from the panelists is not ``UNK’’. So, once the template is selected, we fill in the placeholder with this answer and update $\mathbf{s}^{(t)}$ to generate $q^{(t+1)}$ or the final answer.

\paragraph{Reward}

We use the number of correct answers for CollabQA as \textit{baseline reward}. For each question, getting an correct answer within $T_{max}$ turns leads to a reward $r=+1$; otherwise, the reward $r=-1$.

To alleviate the problem of reward sparsity, we assign the reward $r$ to all the actions in the trajectory. Besides, we add an entropy regularization term to encourage exploration \cite{haarnoja2018soft}. We apply policy gradient method to train $P_0$.  The gradient of the policy is

{\small \begin{align}
    \nabla_{\theta} J  = \mathbb{E}_{\tau \sim \pi}& \left[ \sum_{t=1}^{T} r \nabla_{\theta} \log \pi(a^{(t)} |\mathbf{s}^{(t)} )  \quad + \right.\nonumber\\
                & \left.\nabla_{\theta} \left( \max(0, C - H(\pi(\cdot |\mathbf{s}^{(t)} ))) \right) \right] ~, \label{eq:rl}
\end{align}}%
where $T$ is the turns of dialogue, $C$ is a hyper-parameter, and $\theta$ stands for the parameters of the policy.

In our simple setting, we can introduce an inductive bias specifically tailored to improve the learning effects. Since experts do not share knowledge, and there shall be exactly one response that is not ``UNK'' in each turn, we add an extra negative reward $\beta (\beta <0)$ if it's not the case.
Therefore, the reward $r$ are re-denoted as

{\small\begin{align}
r = \begin{cases}
            -1 + \beta, &\text{if not exactly one answer}, \\
            -1, &\text{if wrong answer}, \\
            +1, &\text{if right answer}.
        \end{cases}
\label{eq:rl_prior}
\end{align}}%
where $\beta$ is a hyper-parameter. As we add prior information in this setting, we call it \textit{enhanced reward} setting.

\section{Experiments and Analysis}
\label{section:experiment}

\subsection{Experimental Setup}

% \paragraph{Settings}

% We train $P_0$ in the following three settings by varying the degree of supervision signal: 1) \textbf{RL from scratch}, $q(t)$ and the responses are withheld, 2) \textbf{RL with warm-up}, where some examples are given. 3) \textbf{Imitation Learning}, or fully supervised, this is the opposite of the first, where $q(t)$ and the responses are always given.

% We train $P_0$ in the following two setting from hard to easy: 1) \textbf{RL} with up to $T$ turns conversation, the reward only comes from the correctness of the final answer, 2) \textbf{RL + prior}, prior here means the number of turns of dialog is known so that conversation always ends after the third turn; and at each turn, a negative reward is given if more than one agents return a non-unknown answer.

\paragraph{Pre-training of the Panelists} We first train the $P_1$, $P_2$, $P_3$ with sub questions and their answers appeared in the training set. We fix the well trained panelists as the environment during training $P_0$. The performance of them are shown in Table~\ref{tab:expert_performance}.

\begin{table}[h]
    \centering
    \begin{tabular}{l|c|c|c}
\hline
         & $P_1$ & $P_2$ & $P_3$  \\
\hline
   Accuracy & 99.6 & 99.6 & 100 \\
\hline
    \end{tabular}
    \caption{Performance of the pre-trained panelists when asked one-hop questions on their domain knowledge.}
    \label{tab:expert_performance}
\end{table}

The hyper-parameters of the model used in our experiments are listed in Table~\ref{tb:hyper}.

\paragraph{Evaluation Metrics} We evaluate the performance of $P_0$ with two metrics: 

1) \textbf{EMA}: exact match of the final answer; 

2) \textbf{EMP}: the extracted reasoning path of $P_0$ exactly matches the ground-truth path.

% The hyper-parameters of the model used in our experiments are listed in Appendix~\ref{appendix:hyper_parameters}.

\begin{table}[t]
\centering
\begin{tabular}{p{10em}|c}
\hline
Hyper-parameter   & Value \\ \hline
R-GCN Layer       & 1     \\
R-GCN hidden size & 80    \\
Embedding dim.    & 40    \\
Bi-LSTM hidden size  & 40    \\
Number of Epoch   & 1000  \\
Batch size        & 500   \\
Optimizer         & Adam  \\
Learning rate     & 3e-3  \\
Entropy threshold $C$ & 0.1   \\
Prior penalty reward $\beta$ & -0.2 \\
\hline
\end{tabular}
\caption{The hyper-parameter used to learn the panelists and $P_0$.}
\label{tb:hyper}
\end{table}

\subsection{Results and error analysis}

\begin{table}[h]
  \centering
  \begin{tabular}{l|c|c}
\hline
                  & EMA & EMP  \\
\hline
Random             & 0.0        & 0.0       \\
Baseline Reward                 & 68.8       & 45.3      \\
Enhanced Reward         & 80.1       & 52.6      \\
% RL w/ 1000 warm-up  & -     & -     \\
% Supervised                  & -       & -       \\
\hline
  \end{tabular}
  \caption{Main results of the experiments on the test set.}
  \label{tab:main_results}
\end{table}

\begin{figure}[t]
\centering
\includegraphics[width=0.5\textwidth]{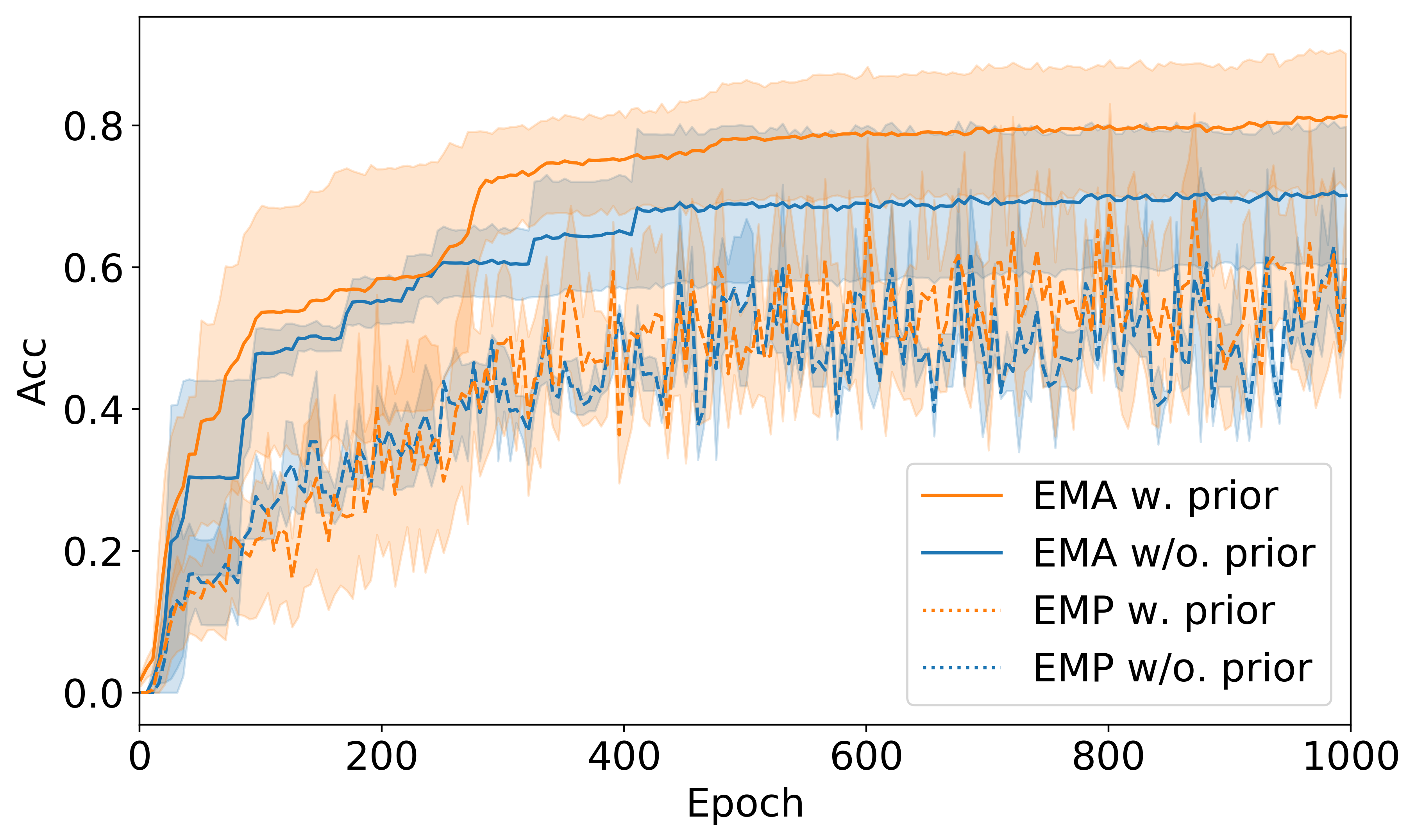}
\caption{The training curve for $P_0$ with baseline and enhanced reward . The final answer accuracy (EMA) will tend to converge, while the reasoning path accuracy (EMP) fluctuates. Adding prior will make the training faster and achieve better final accuracy, but cannot reduce the reasoning path accuracy fluctuation. }
\label{fig:training_curve}
\end{figure}

The main results are shown in Table~\ref{tab:main_results}. In the top row, we show the performance for a random $P_0$ who randomly picks one question to ask at each step. Its EMA is zero, which reveals that it is not easy to guess the right answer. The ``baseline reward'' row presents the results of using the Equation~\ref{eq:rl} as the gradient to optimize the model, and this model has one more termination action at each step (and it should pose the ``termination'' action only at the 4th turn.). The ``enhanced reward'' row is the additional penalty we give in Equation~\ref{eq:rl_prior} (i.e. knowing exact number of turns and only one response is not ``UNK'').
% Even though the total action space is limited (each step selects among 28 actions, leading to a total action space of $28^3 = 21,952$), this task is still challenging.

\paragraph{Performance gap between two kinds of rewards} The accuracy difference between the ``baseline reward'' row and the ``enhanced reward'' row is caused by that the ``baseline reward'' setting has one extra action, therefore it has the chance to terminate too early or fail to stop at the last turn. Based on our experiments, we found that this improper termination accounts for near 9\% of the total errors. However, this kind error can be totally avoided in the ``enhanced reward'' setting. The left 2.3\% performance lost may attribute to better training of ``enhanced reward''. Another noticeable fact is that the EMP drop from ``enhanced reward'' to ``baseline reward'' is not as large as the EMA, this is because the wrong reasoning path in ``enhanced reward'' is also prone to terminate improperly. The training curves for ``baseline reward'' and ``enhanced reward'' are presented in Figure~\ref{fig:training_curve}, the confidence interval is calculated from 5 experiments. From the figure, the accuracy for the EMA is quite stable, while the EMP is fluctuating. This is because the number of distinct samples are not very rich in our dataset, the variance should be innately small. However, because of the data bias which will be discussed in the following part, the EMP will fluctuate and without hurting the EMA.

% We note that adding prior improves the performance of both answer accuracy and reasoning path accuracy. It shows that knowing the exact number of hops of the question and only one expert can answer the sub question at each turn, makes the task easier in our dataset. These priors are generally unavailable in the real world applications.

\paragraph{Fitting the data bias} What is interesting is that a high answer accuracy (EMA) does not mean a high reasoning path accuracy (EMP); in both settings there is a large gap between the two accuracies.

\begin{figure}[t]
\centering
\includegraphics[width=0.4\textwidth]{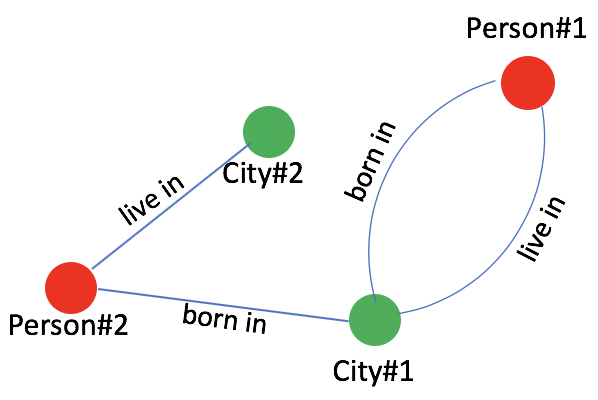}
\caption{One example about the data bias in our data, 99\% of persons have the same ``live in'' and ``born in'' city. }
\label{fig:bias}
\end{figure}

To understand the reason behind the gap between answer accuracy and reasoning path accuracy, we compute the performance for each type of $Q$. We found that the model finds wrong reasoning path mostly on the questions that has sub questions ``Which city does [PersonName] live in ?'' and ``Which city was [PersonName] born in ?''. It turns out that, in our dataset, nearly 99\% of the time that a person's ``birthplace'' and ``live in place'' are the same, and the model cannot distinguish the difference between them during training. We observe that nearly all the questions that need to be decomposed to ``Which city does [PersonName] live in ?'' have been decomposed to ``Which city was [PersonName] born in ?'' instead. One example can be viewed in Figure~\ref{fig:bias}.

\begin{figure}[t]
\centering
\includegraphics[width=0.4\textwidth]{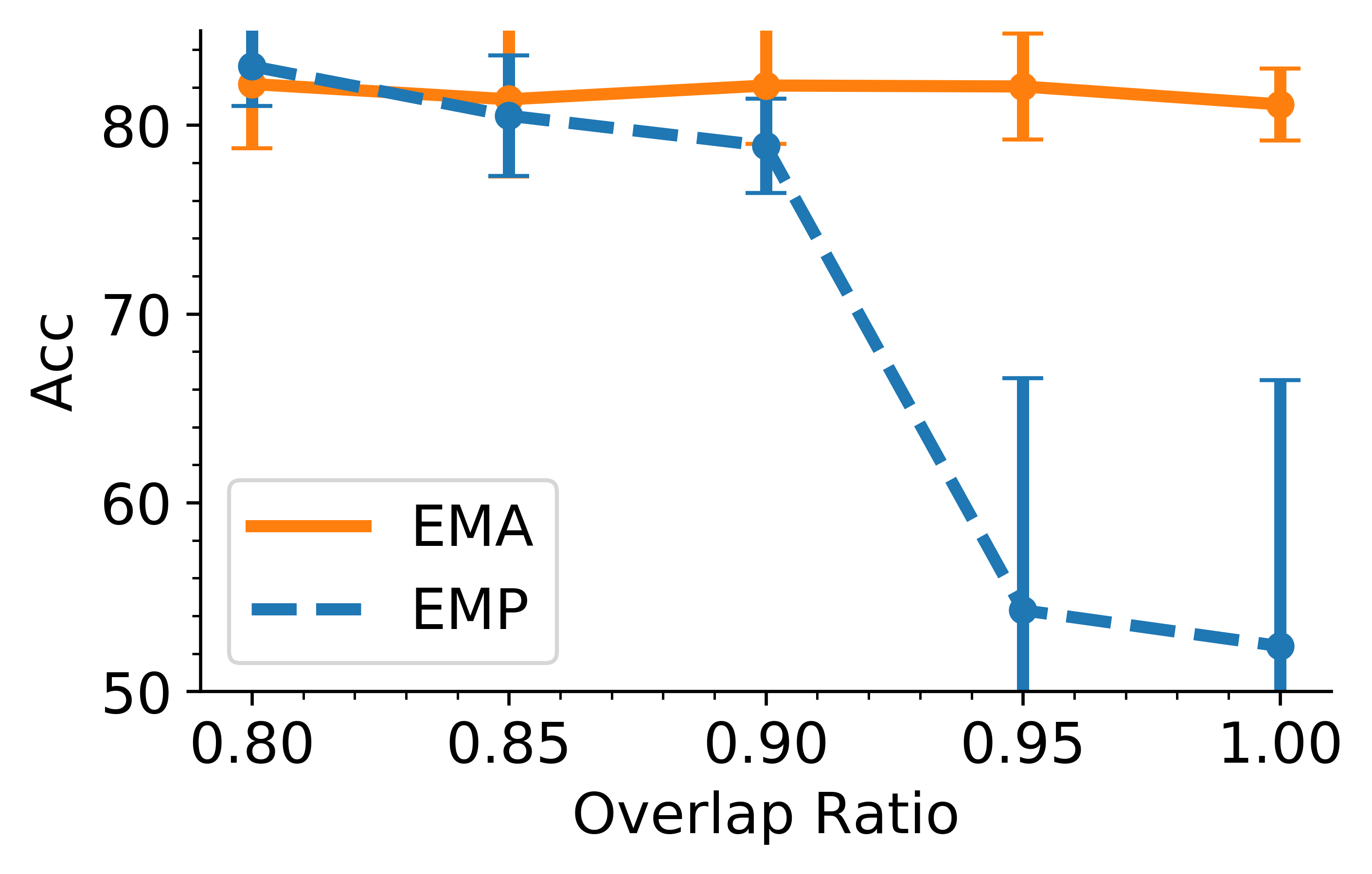}
\caption{The final answer accuracy and reasoning path accuracy with respect to the overlap ratio. }
\label{fig:overlap}
\end{figure}

To further show how this overlap will have an impact on our results,  we vary the overlap ratio, which is the probability one person has the same ``live in placce'' and ``birthplace'', in Figure~\ref{fig:overlap}. As the overlap ratio goes up, it will be harder for P0 to discern between ``live in placce'' and ``birthplace'', but the difficulty does not go up linearly, the EMP drops sharply after some point. However, the the drop of EMP does not have too much negative affect on the EMA, since when the overlap ratio is high, it can still get the right answer with the wrong reasoning path.

In other words, the model settles on an approximate question decomposition that the final reward cannot distinguish.  We note that similar data bias exist in the real world, and the model exploited it through exploration in our dataset. Fixing it means $P_0$ should inspect the semantic consistency between the sub questions and the original question, instead of blindly selecting templates.

Except for the data bias issue, there are numerous error cases. For instance, an imperfect an expert can pick a wrong answer that are structurally correct (i.e. answer the birth city when the question is work location) that leads to a correct decomposition but wrong final answer. Note that our panelists are nearly perfect. Thus, small errors accumulate over the turns and greatly affect $P_0$'s performance.

%Note that this is not bug but a feature, for there are similar data bias in the real world, and the model figured it out through exploration in our dataset.
% To explore the reason behind the gap between answer accuracy and reasoning path accuracy, we compute the performance for each type of $Q$. We found that the model finds wrong reasoning path mostly on the questions that has sub questions ``Which city does [PersonName] live in ?'' and ``Which city was [PersonName] born in ?''. It turns out that, in our dataset, nearly 99\% of the time that a person's ``birthplace'' and ``live in place'' are the same, and the model cannot distinguish the difference between them during training. We observe that nearly all the questions that need to be decomposed to ``Which city does [PersonName] live in ?'' have been decomposed to ``Which city was [PersonName] born in ?'' instead. Note that this is not bug but a feature, for there are similar data bias in the real world, and the model figured it out through exploration in our dataset.

\paragraph{The problem of group bias}

The data bias described above reminds us of another kind of bias, that during learning to collaborate, $P_0$ may fit the bias of the panelists.
%The bias of the panelist means, panelist trained with different initialization may have different behaviors, for they may fit different data bias in the dataset.
This is intuitive, since panelists are environment, any bias therein will lead to bias in $P_0$.
What is more interesting in a collaboration setting is that such bias is per \emph{group}. %So $P_0$ trained with different group of panelist may have been taught with different ``knowledge''.

To verify this assumption, we conduct experiments training 3 groups of panelists with different initialization, and we call them $Panel^{(1)}, Panel^{(2)}, Panel^{(3)}$. Each group has similar performance to those in Table~\ref{tab:expert_performance}. Then we train 3 versions of $P_0$ paired with different groups of panelist: $P_0^{(i)}$ trains with $Panel^{(i)}$. During testing, we pair each $P_0$ with different group of panelist. The the resulting answer accuracy are listed in Table~\ref{tab:group_bias}.

\begin{table}[h]
    \centering
    \begin{tabular}{l|c|c|c}
\hline
                & $P_0^{(1)}$   & $P_0^{(2)}$ & $P_0^{(3)}$ \\
\hline
$Panel^{(1)}$    & \textbf{81.7}             & 82.5              & 76.4              \\
$Panel^{(2)}$     & 80.0             & \textbf{84.6}             &  74.2             \\
$Panel^{(3)}$     & 80.8              & 83.6              & \textbf{81.9}              \\
\hline
    \end{tabular}
    \caption{Results for pairing each $P_0$ with different group of panelist at testing time. Each column shows how a version of $P_0$ is paired with different panels; the diagonal entry is where $P_0$ pairs with the group it was trained on.}
    \label{tab:group_bias}
\end{table}

The results show that there are always performance drop when $P_0$ is paired with panelists which is not trained with it.

% \textcolor{red}{Evidence to show different panelist learns different bias...}

% \begin{table}[h!]
% \centering
%     \begin{tabular}{l|c}
%     \toprule
%                                   & Acc / EM  \\
%     \midrule
%     $A_0$ Decomposition EM w/ GT    & 100  \\
%     $A_1$ Answering Acc w/ GT    & 99.2 \\
%     $A_2$ Answering Acc w/ GT    & 100  \\
%     $A_3$ Answering Acc w/ GT    & 100  \\
%     Joint Answering Acc w/o GT   & 97.7 \\
%     \bottomrule
%     \end{tabular}
%     \caption{Supervised training results. }
% \end{table}

\section{Related work}
\label{section:related_work}

\paragraph{KGQA}

In the simplified setting of CollabQA, each panelist is a simple KGQA system. The questions are either simple, or need one step reasoning to be reformulated into another simple question.

KGQA has been widely studied. The most common way of doing KGQA is by semantic parsing. Semantic parser maps a natural language question to a formal query such as SPARQL, $\lambda$-DCS \cite{liang-etal-2011-learning} or FunQL \cite{liang-etal-2011-learning}. Previous works on KGQA can be categorized into classification based, ranking based and translation based methods \cite{chakraborty2019introduction}. The model of the panelists we proposed is most related to classification based methods. Classification based methods assume the target formal query has a fixed structure, and the task is to predict the elements in it. For example, in SimpleQuestions benchmark \cite{bordes2015large}, all the questions are factoid questions that need one-step reasoning. SimpleQuestions has been approached by various NN models \cite{he-golub-2016-character,dai-etal-2016-cfo,yin-etal-2016-simple,yu-etal-2017-improved,10.1145/3038912.3052675,mohammed-etal-2018-strong,petrochuk-zettlemoyer-2018-simplequestions,10.1145/3289600.3290956}.

Another line of KGQA approaches leverages knowledge graph embedding to make full use of the structural information of KGs~\cite{10.1145/3289600.3290956}.

\paragraph{Multi-hop QA}
%Recently, multi-hop QA attracts a lot of attention in NLP community.
%The questions that CollabQA aims to solve are mostly multi-hopped.
To answer a multi-hopped question, multiple supporting facts are needed. WikiHop \cite{welbl-etal-2018-constructing} and HotpotQA \cite{yang-etal-2018-hotpotqa} are recently proposed multi-hop QA datasets for text understanding.
%Though in the two tasks, the supporting facts are scattered in different documents, they are accessible to the model.
Different from multi-hop QA, the supporting facts in CollabQA are separately owned by different panelists. Each supporting fact is not accessible except its owner. Therefore, CollabQA is more challenging than multi-hop QA. %Recent works approach multi-hop QA by decomposing the original complex question into several simple questions \cite{DBLP:conf/acl/MinZZH19,DBLP:journals/corr/abs-2001-11770,DBLP:journals/corr/abs-2002-09758}, which is similar to CollabQA. The key difference is that the decomposition in these works depends on the question itself, while in CollabQA, the question is not necessarily complex, and the ``decomposition" is dependent on the KG of the agent.

%\subsection{Question Generation}
%The questions generated by agents need to be natural languages, thus it shares similarity with the question generation task, which has been defined as ``automatically generating questions from various inputs such as raw text, database, or semantic representation.'' in \cite{rus2008}. Since the introduction of neural networks, various question generation scenarios have been evaluated. \citet{DBLP:conf/acl/SerbanGGACCB16,DBLP:conf/naacl/ElSaharGL18} built models to generate questions based on the SimpleQuestion dataset \cite{DBLP:journals/corr/BordesUCW15}, the model is provided with a knowledge base triple, and it should be able to ask a question based on this triple. Given the passage and answer from the SQuAD dataset \cite{DBLP:conf/emnlp/RajpurkarZLL16}, \citet{DBLP:conf/acl/DuSC17,DBLP:journals/corr/abs-1908-04942} managed to generate a coherent question. \citet{DBLP:conf/acl/MostafazadehMDM16,DBLP:conf/emnlp/Faruqui018} tried to generate questions based on pictures and ill-formed questions, respectively. Reinforcement learning is also used in question generations to generate more coherent utterance \cite{DBLP:journals/corr/abs-1907-09899,DBLP:journals/corr/abs-1908-04942}.

\paragraph{Multi-Agent Reinforcement Learning (MARL)}
In this paper, panelists are passive and pre-trained, we just train the collaborative policy under single-agent RL setting. However, the general CollabQA should allow the panelists discuss with each other; therefore, each panelist has its own policy and can update the policy. Under this general setting, CollabQA naturally falls into the realm of MARL~\cite{bucsoniu2010multi,foerster2016learning}, which is a more challenging task.

\section{Discussion}
\label{section:discussion}
The task of CollabQA as it stands is very simple. Nevertheless, the experiences are helpful to drive towards an improved setting that is closer to real-world scenarios. To put it differently, if we were to design the task anew, what are the most important extensions? We examine three dimension: 1) the role and capability of participants, 2) the collaboration structure and 3) scaling to real-world problems.

% This is obviously a difficult question. We identify the following four dimensions: 1) dialogue structure; 2) KG sharing 3) ground truth to be discovered and 4) natural language question and answer generation. We now discuss each in turn.
% We identify the following gaps: 1) relaxing the current dialogue structure which pairs of an active moderator and a passive panel; 2) removing the share-nothing assumption among the participants; 3) enriching the nature of the knowledge fragment that the system aims to recover and finally 4) lifting the constraint of using template to select questions. We now discuss these gaps in more details.

\paragraph{(1) Role Definition}

In the current setting, the moderator $P_0$ assumes no knowledge of its own and its capacity is limited to breaking down a complex question. The panelists are domain experts whose knowledge do not overlap, and they can only respond with facts, without proactively ask questions, nor can they reveal any reasoning path. These are much simplified assumptions that do not reflect the reality. Relaxing these constraints is in general challenging; we list some of the issues below.

Consider the issue of \emph{common sense knowledge}. Although inconsistencies among individuals do exist, it is nevertheless the foundation where collaboration among a collection of human experts can start. Often times, common sense is required to meaningfully decompose a complex question, whether the panelists are involved or not.
Take the question ``Does Person\#1 work in the same city as Person\#2 ?'' as an example. $P_0$ needs to realize that the entities of the companies and their locations are key to solve this question. These missing steps, which are not obvious from the question itself, need to be inserted and it takes common sense to deduce them, since ``working city'' is not a relation readily available in our KG.

A debate is interesting when there are gaps between experts, not because they have non-overlapping knowledge but more often because they have different opinions on the same facts. As such we need to introduce \emph{overlapping knowledge} imbued with different certainty (or reliability). This, in turn, requires $P_0$ to have the capacity to arbitrate among parallel responses from different panelists.

\paragraph{(2) Collaboration Structure}
The overall structure of a moderator working together with a group of expert is not uncommon. Even with this broad structure, there can be other valid variations. For instance, instead of broadcast, the moderator can have pointed question to one panelist, or more generally a subset of the panelists. It is also possible that the final response needs a vote when the moderator cannot resolve a difference.

The constraint that panelists can only passively state facts is problematic when a question is ambiguous. Consider the question ``Where does [PersonName] work ?'' There are multiple legitimate responses (e.g. a company, a city, and/or a country). As such, a panelist should ask \emph{clarification question}; drawing an exhaustive list from KG is a possibility, but an unnatural one. As a further extension, clarification questions can be generated and responded by any of the participants.

\paragraph{(3) Scale to Real World Scenario}
Despite its simplicity, our current setting is meaningful to approach real CollabQA tasks. In order to do so we believe there are few more necessary extensions.

%\subparagraph{The Many Roads that Lead to Roman}
Currently, we assume a complex question is the realization of a unique path. In general this is not true even when the reasoning does take a multi-hop path; multiple edges can exist between a pair of entities. A lazy (or unlucky) $P_0$ may learn to choose only one of them, if the only award is to get the final answer right. This is one problem we discussed in our experiments where ``work\_in'' and ``live\_in'' happen to overlap in the end nodes.

In general, \emph{reasoning can take a graph} (thought we can consider a path as a degenerated graph, too). The earlier example (``Does Person\#1 work in the same city as Person\#2 ?'') can only be solved by a two-level tree with a Boolean comparison at the root. Booking an airline ticket with both pricing and timing constraints while the required information reside in different KGs is similar. As a result, to generate complex question there is a need to go beyond the perspective of a reasoning path.

%\subparagraph{Natural Language Question and Answer Generation}
In our current setting, $P_0$ selects template, and panelists respond with the entity. As such, the action space of $P_0$ is constrained, and there is very low risk that communications get ``lost in translation.'' Ideally, such communication should take generated natural language. In other words, CollabQA needs \emph{natural language generation} (NLG) as a component. However, doing so will be prohibitive expensive if we want to train from scratch. If there are $|V|$ valid words, the number of possible sentences for a $L$ length sequence will be $|V|^{L}$, and that is only for one turn. This will exponentially exacerbate the issue of sparse reward, making training difficult. Thus, we believe that this is not a fundamental problem. In other words, in the context of CollabQA, leaning what to ask is more important than how to ask. A more practical approach is using transfer learning to endorse the agents with NLG capability.

However, there should be surface realization diversity even for semantically identically questions. Doing so is not only a practically required, but will also make the system more robust. This can be easily accomplished by adding noises to templates, provided that the action space stays manageable.

\section{Conclusion}
The fact that knowledge are not shared gives rise to individual diversity and motivates collaboration. We believe natural-language based collaboration system is a domain that has practical implication and holds scientific values. The CollabQA task and dataset we proposed in this paper is a small step towards that direction.

\bibliography{emnlp2020}
\bibliographystyle{acl_natbib}

\newpage
\clearpage

\appendix
\appendixpage
\addappheadtotoc

\section{Details of the CollabQA dataset}
% \subsection{Knowledge Graphs}
\label{appendix:kg_details}

\begin{figure}[t]
\centering
  \begin{subfigure}{0.5\textwidth}
    \includegraphics[width=\linewidth]{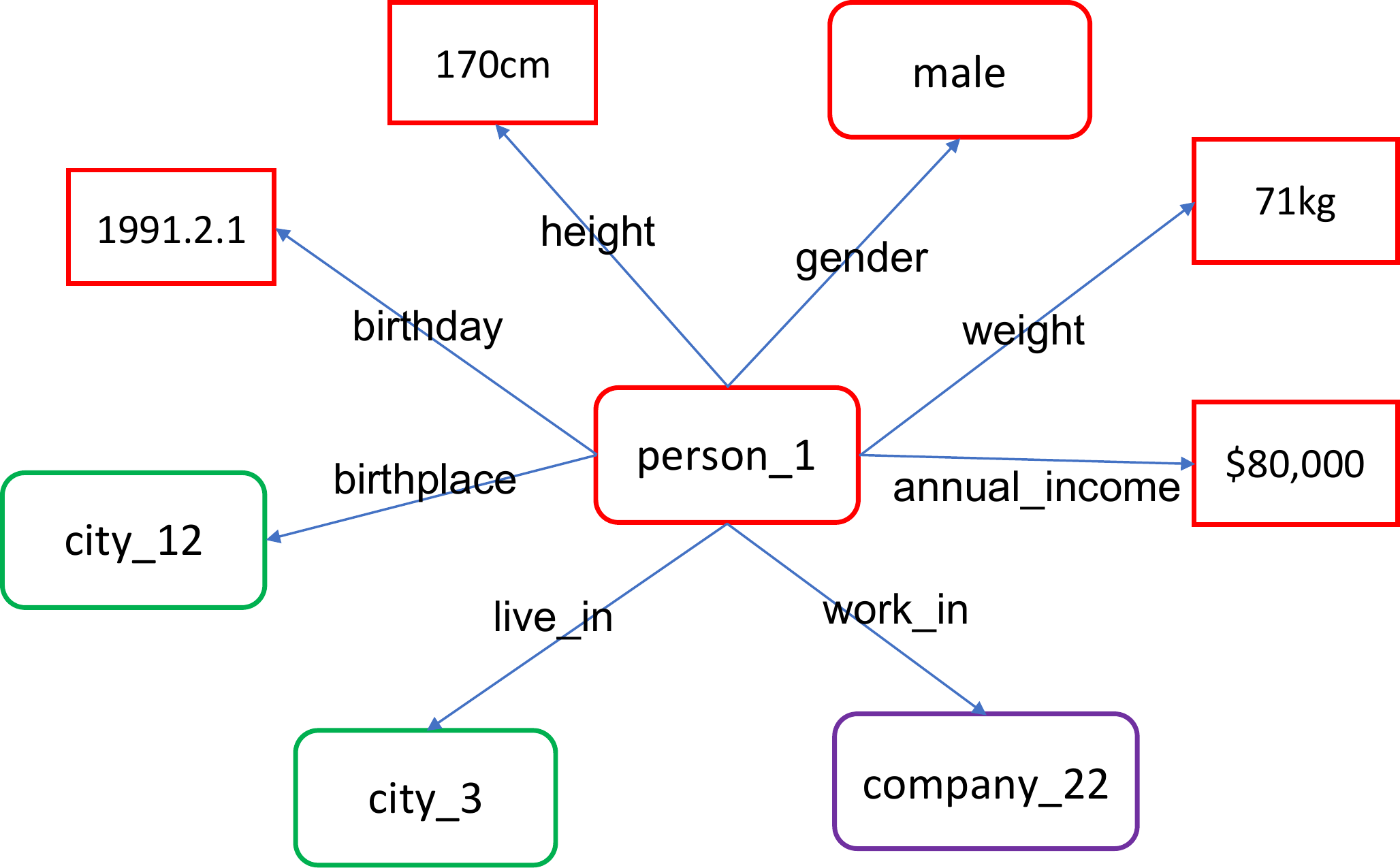}
    \caption{An example of Person entity in $G_1$} \label{fig:g1}
  \end{subfigure}
  \vspace{0.5cm}
  \begin{subfigure}{0.5\textwidth}
    \includegraphics[width=\linewidth]{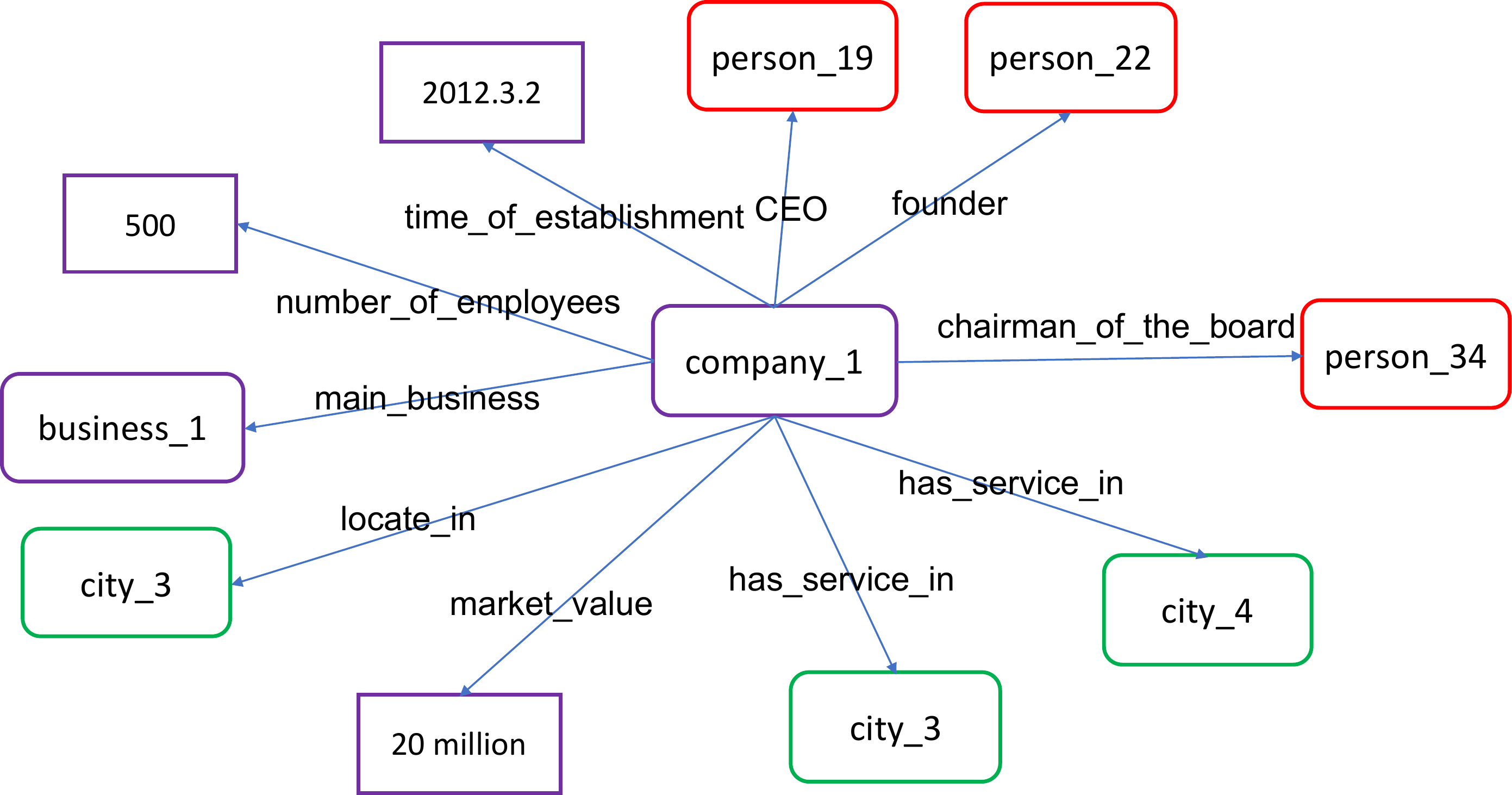}
    \caption{An example of Company entity in $G_2$} \label{fig:g2}
  \end{subfigure}
  \vspace{0.5cm}
  \begin{subfigure}{0.5\textwidth}
    \includegraphics[width=\linewidth]{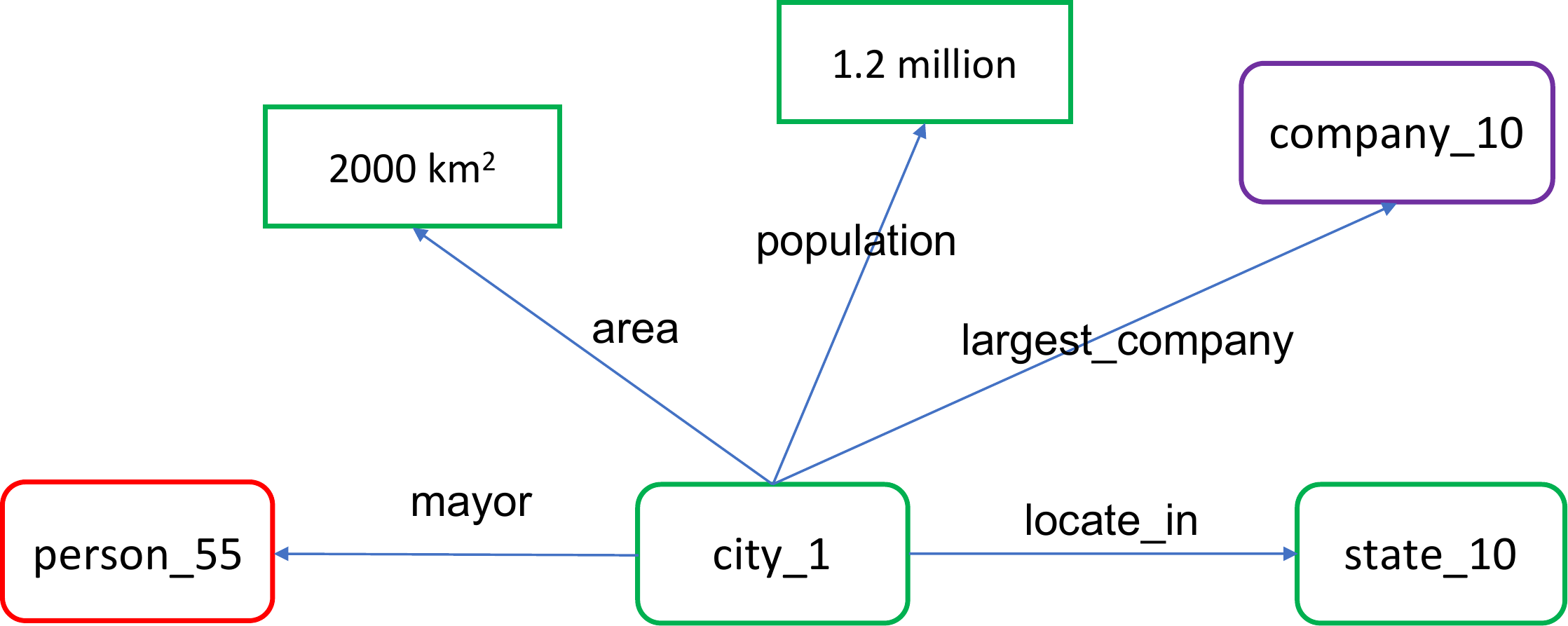}
    \caption{An example of City entity in $G_3$} \label{fig:g3}
  \end{subfigure}

\caption{Structure and examples of entities in the three proposed knowledge graphs. } \label{fig:kg}
\end{figure}

\paragraph{Structures of the three KGs}
Figure~\ref{fig:kg} shows the structure and examples in our proposed knowledge graphs. $G_1$ contains a list of \textit{Person} entities. The value of a property of the entity is randomly generated within a reasonable range. For example, the value of a person's height is randomly sampled in the range $[160cm, 200cm]$. We add a series of constraints to make the KGs more realistic, such as a person who doesn't have job gets no annual income; a person cannot be a mayor and be an employee in some company at the same time; the largest company of a city must be located in that city, and so on.

\paragraph{Statistics of the KGs}
The detailed statistics of the three KGs are shown in Table~\ref{tab:kg_statistics}.

\begin{table*}[t]
\centering
\begin{tabular}{|l|l|l|l|}
\hline
                                                    & \multicolumn{1}{c|}{$G_1$}   & \multicolumn{1}{c|}{$G_2$} & \multicolumn{1}{c|}{$G_3$} \\ \hline
\multirow{2}{6em}{Overall}                            & Number of entities: 7541    & Number of entities: 7719   & Number of entities: 1360    \\
                                                    & Number of relations: 24000  & Number of relations: 16000 & Number of relations: 1500   \\ \hline
\multirow{8}{6em}{Number of different node types}     & gender\_value: 2             & CompanyName: 2000          & CityName: 300              \\
                                                    & PersonName: 3000            & date\_value: 1862           & area\_value: 211            \\
                                                    & height\_value: 21            & number\_value: 836         & number\_value: 259          \\
                                                    & weight\_value: 31            & PersonName: 2600           & PersonName: 285            \\
                                                    & date\_value: 2597           & BusinessName: 20           & CompanyName: 300           \\
                                                    & CityName: 300                & CityName: 300              & StateName: 5               \\
                                                    & CompanyName: 1559            & market\_value: 101         &                            \\
                                                    & annual\_income\_value: 31 &                            &                            \\ \hline
\multirow{9}{6em}{Number of different relation types} & height: 3000                & establish\_date: 2000      & area: 300                  \\
                                                    & weight: 3000                & number\_of\_employees:2000 & population: 300            \\
                                                    & birthday: 3000              & ceo:2000                   & mayor: 300                 \\
                                                    & gender: 3000                & founder:2000               & largest\_company: 300      \\
                                                    & birthplace: 3000            & main\_business:2000        & contained\_by: 300         \\
                                                    & live\_in: 3000              & locate\_in:2000            &                            \\
                                                    & work\_in: 3000              & has\_service\_in:2000      &                            \\
                                                    & annual\_income: 3000        & chairman:2000              &                            \\
                                                    &                              & market\_value:2000         &                            \\ \hline
\end{tabular}
    \caption{Statistics of three knowledge graphs used in our experiment.}
    \label{tab:kg_statistics}
\end{table*}

\end{document}